\documentclass[11pt]{article}

\usepackage{dialogue}
\usepackage{xspace}
\usepackage{ifxetex}
\ifxetex
    \usepackage{fontspec}
    \setromanfont{Times New Roman}
\else
  \usepackage{cmap}
  \usepackage[T2A,T1]{fontenc}
  \usepackage[utf8]{inputenc}
  \usepackage{times}
  \usepackage{latexsym}
  \usepackage{substitutefont}
  \substitutefont{T2A}{\familydefault}{cmr}
\fi
\usepackage{tabularx}
\usepackage[russian,british]{babel}
\usepackage{url}
\usepackage{pgf}

\usepackage{covington} 

\usepackage{hyperref}

\dialogfinalcopy 

\title{DIALOG-22 RuATD Generated Text Detection}
\author{
  Narek Maloyan \\
  Lomonosov MSU \\
  {\tt maloyan.narek@gmail.com} \\\And
  Bulat Nutfullin \\
  Lomonosov MSU \\
  {\tt bulat15g@gmail.com} \\\And
  Eugene Ilyushin \\
  Lomonosov MSU \\
  {\tt john.ilyushin@gmail.com} \\
}

\date{}

\begin{document}
\maketitle

%
%
%
%
\begin{abstract}
  Text Generation Models (TGMs) succeed in creating text that matches human language style reasonably well. Detectors that can distinguish between TGM-generated text and human-written ones play an important role in preventing abuse of TGM.
  In this paper, we describe our pipeline for the two DIALOG-22 RuATD tasks: detecting generated text (binary task) and classification of which model was used to generate text (multiclass task) \cite{shamardina2022ruatd}. We achieved 1st place on the binary classification task with an accuracy score of 0.82995 on the private test set and 4th place on the multiclass classification task with an accuracy score of 0.62856 on the private test set. We proposed an ensemble method of different pre-trained models based on the attention mechanism\footnote{\href{github.com/maloyan/ruatd}{github.com/maloyan/ruatd}}.

  \textbf{Keywords:} Generated text, text classification, ensemble methods, multi-class classification task

\end{abstract}

%
%
%
%
\selectlanguage{russian}

  

%
%
%
%

\begin{abstract}
Модели генерации текста успешно синтезируют текст, который сложно отличим от написанного человеком текста. Детекторы, способные отличить текст, созданный автоматически, от написанного человеком текста позволяют предотвратить злоупотребление сгенерированными текстами.
В этой статье мы описываем наше решение для задач DIALOG-22 RuATD по обнаружению сгенерированного текста и классификации с помощью какой модели был сгенерирован текст. 
Мы заняли 1-е место в задаче бинарной классификации с оценкой точности 0,829 в частном наборе тестов и 4-е место в задаче мультиклассовой классификации с оценкой точности 0,628 в  тестовом наборе.
Наше решение является ансамблем дообученных моделей, основанных на механизме внимания.

  \textbf{Ключевые слова:}  Сгенерированные тексты, классификация текстов, ансамблевые методы, задача многоклассовой классификации
\end{abstract}
\selectlanguage{british}

%
%
%
%
\section{Introduction}
\label{intro}
%
%

As the language neural nets got better at generating texts, it's getting harder and harder to distinguish the human-written text from generated one. So manual detection of these texts got impossible. For that reason, it's desirable to build a system that can automatically detect generated text. 

The proposed system will use various features extracted from the text such as length, punctuation, word choice etc. To determine whether the text is human-generated or not. The accuracy of this system can be improved by using machine learning algorithms which will learn how humans generate texts and then use those features for detection purpose.

There are many ways to build such a system, but probably the most reliable one is based on machine learning algorithms. These algorithms can be trained on a large number of examples - both human-generated and computer-generated texts. After being trained, they should be able to identify which texts are computer-generated with high accuracy.

This approach already works in other areas, such as spam detection. Some early experiments have shown promising results and indicate that this approach works well for the detection of the generated text.

The DIALOG-22 shared task of RuATD 2022 had 2 tracks for binary classification and multiclass classification. In this report, we will describe the data, how we handled it, the models we used, and the ensembling technique.

%
%
%
%
\section{Task Definition}
For the binary classification task $F_{binary}$, we frame the generated text detection problem as follows: given a piece of text $X$, label it as either human-written or machine-generated $y_{binary} = \{H,M\}$.

$$F_{binary}: X \rightarrow y_{binary}$$

For the multiclass classification task $F_{multiclass}$, we set up the problem as follows: given a piece of text $X$, label it as one of the 14 classes that represent deep neural models 
$y_{multiclass}$ = \{ 
    \textit{
        M2M-100, Human, OPUS-MT, M-BART50, ruGPT3-Medium, ruGPT3-Small, 
        mT5-Large, ruGPT3-Large, ruT5-Base-Multitask, mT5-Small, 
        ruT5-Base, ruGPT2-Large, M-BART, ruT5-Large
    }
\}

$$F_{multiclass}: X \rightarrow y_{multiclass}$$

%
%
%
%
\section{Datasets}
Provided datasets offer the train and test splits. The part of the set was annotated automatically by different generative models. Various language models were fine-tuned on different tasks: machine translation, paraphrasing, summarization, simplification, and unconditional text generation - are used to generate texts. The texts written by a human were collected from open sources from different domains. \cite{ruatd-2022-binary}, \cite{ruatd-2022-multiclass}.

\begin{table}[h!]
\selectlanguage{russian}
  \begin{center}
    \begin{tabular}{|l|c|}
      \hline \bf Text & \bf Class\\ \hline
        Обустройство тротуаров, мостовых (в том числе тротуарной плиткой).           & H              \\
        Минстрой обозначил способы снижения энергоемкости российской экономики.      & M              \\
        В конце 1873 года военный суд вынес решение по делу Франциска Ахиллы Базейн. & M              \\
        увеличение правовой грамотности и развитие правосознания граждан.            & H              \\
      \hline
    \end{tabular}
  \end{center}
  \selectlanguage{british}
  
  \caption{Example of the binary classification data.}
  \label{tabFonts_binary}
\end{table}

\begin{table}[h!]
\selectlanguage{russian}
  \begin{center}
    \begin{tabular}{|l|c|}
      \hline \bf Text & \bf Class\\ \hline
        Прочла автобиографию Каутского, Одесса, 1905. & Human           \\
        Вы не можете быть в печи и в муulin.          & M-BART50        \\
        Вторая попытка привела к тому же результату.  & OPUS-MT         \\
        Сколько учеников в вашем классе?              & M2M-100         \\
      \hline
    \end{tabular}
  \end{center}
  \selectlanguage{british}
  \caption{Example of the multiclass classification data.}
  \label{tabFonts_multiclass}
\end{table}

%
%
%
%
\section{Related Works}
In this section, we will discuss various methods for detecting machine-generated texts.

Over the past years, many approaches appeared for generated text detection. The latest works are usually based on using transformer-based models. Either fine-tuning the proposed task or using probabilities distribution to make decision-based on them \cite{ippolito2019automatic}. Here we list some examples of different methods:

\begin{itemize}
    \item First, we calculate the mean likelihood over all machine-generated sequences, then the mean of human-written ones. If the likelihood according to some language model is closer to the machine-generated mean likelihood, then we consider it as generated text \cite{solaiman2019release};
    \item In GLTR \cite{gehrmann2019gltr} described a method using a language model to compute the probability distribution of the next word given the text sequence. For each sequence position, we get the likelihood of the ground-truth next word within this list. Then these
ranks are displayed on a histogram. Based on the distribution of bins, we can decide if this text is generated or not.
    \item Bert fine-tuning on the classification task. Having a label of text if it's machine-generated or not, we can fine-tune the language model to predict \cite{solaiman2019release}
    \item Also, there is possible to use human-machine collaboration. Real or Fake Tool provides a game-like interface for humans to decide at what point the text begins to be written by a computer \cite{dugan2020roft}. 
\end{itemize}

More approaches are described in the survey \cite{jawahar2020automatic}.

%
%
%
%
\section{Experiment Setup}

In this section, we present the experiment configurations we use to solve binary and multiclass tasks.

\subsection{Data preproccessing}

We decided not to perform any preprocessing of the text itself. Regarding the data split:

\begin{itemize}
    \item \textbf{Binary classification}. We concatenated the train set with the validation set. This new concatenated dataset gave us an option to perform a 5-fold cross-validation; 
    \item \textbf{Multiclass classification}. We took the data as is without changing anything.
\end{itemize}

\subsection{Models}
All models we used for fine-tuning we got from transformers library \cite{wolf2020transformers}. Here we will describe these models:
\begin{itemize}
    \item \textbf{sberbank-ai/sbert\_large\_nlu\_ru}\cite{habr_2020}, \textbf{sberbank-ai/ruBert-large}, \textbf{DeepPavlov/rubert-base-cased} are BERT models \cite{devlin2018bert}. Sber-ai models were fine-tuned on a closed dataset collected by their research group \cite{sberbank-ai}. DeepPavlov's RuBERT was fine-tuned on the Russian part of Wikipedia and news data.
    \item \textbf{IlyaGusev/mbart\_ru\_sum\_gazeta} is a mBART model fine-tuned on the dataset for summarization of Russian news\cite{gusev2020dataset}. The original MBart model was pretrained on large-scale monolingual corpora in many languages using the BART objective. mBART is one of the first methods for pre-training a complete sequence-to-sequence model by denoising full texts in multiple languages, while previous approaches have focused only on the encoder, decoder, or reconstructing parts of the text \cite{liu2020multilingual}.
    \item \textbf{MoritzLaurer/mDeBERTa-v3-base-mnli-xnli}  \cite{he2021debertav3} This multilingual model is suitable for multilingual zero-shot classification. The original model was pre-trained by Microsoft on the CC100 multilingual dataset \cite{wenzek-etal-2020-ccnet}.
    
    This model was fine-tuned on the XNLI development dataset and the MNLI train dataset. The XNLI development set consists of translated texts for each of the 15 languages \cite{conneau2018xnli}.

    \item \textbf{DeepPavlov/xlm-roberta-large-en-ru-mnli} is an XLM-RoBERTa model \cite{conneau2019unsupervised} which was fine-tuned on the english-russian part of the MNLI \cite{williams2017broad} dataset.
\end{itemize}

\subsection{Binary classification (with ensembling technique)}
Five chosen models were used in the experiement: \textbf{sberbank-ai/sbert\_large\_nlu\_ru}, \textbf{sberbank-ai/ruBert-large}, \textbf{IlyaGusev/mbart\_ru\_sum\_gazeta }, \textbf{MoritzLaurer/mDeBERTa-v3-base-mnli-xnli},  \textbf{DeepPavlov/xlm-roberta-large-en-ru-mnli}

\begin{figure}[h!]
 \centering
  \includegraphics[width=0.6\textwidth]{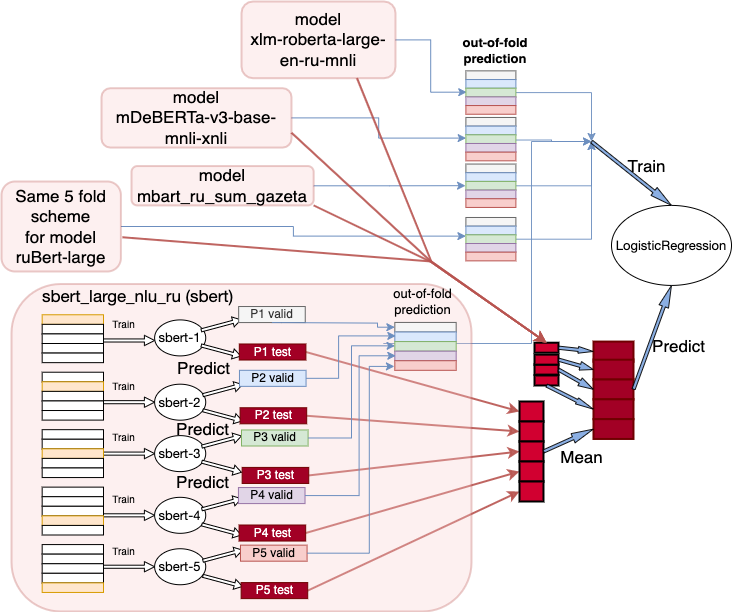}
  \caption{Ensembling scheme}
  \label{fig:architcture}
\end{figure}

Let's describe steps for training each of these models:
\begin{enumerate}
    \item We split our training dataset into non-overlapping 5-folds and performed cross-validation (Figure \ref{fig:architcture});
    \item for each validation fold we predicted the target and as a result, we got out-of-fold predictions for training data;
    \item for the test set, we predicted the target from every 5 models from cross-validation and averaged the result which is marked with red;
\end{enumerate}

As a meta-model, we choose Logistic Regression, which was trained on out-of-fold predictions. Then we predict the final results for the test set.

\subsection{Multiclass classification}
For the multiclass classification we chose these models: \textbf{DeepPavlov/rubert-base-cased}, \textbf{DeepPavlov/xlm-roberta-large-en-ru}, 
\textbf{IlyaGusev/mbart\_ru\_sum\_gazeta}. We fine-tune these models without cross-validation only on provided train set.

\section{Results}

Table \ref{binaryTable} shows each model accuracy score on binary classification task. The multiclass classification accuracy is shown in Table \ref{multiclass_classification}. On both tasks, the best performing single model was \textbf{DeepPavlov/xlm-roberta-large-en-ru-mnli}. We managed to ensemble models in the binary task, so the ensemble of models showed the best accuracy.

\begin{table}[h!]
  \begin{center}
    \begin{tabular}{|l|c|}
      \hline \bf Model name & \bf Accuracy\\ \hline
        sberbank-ai/sbert\_large\_nlu\_ru                  & 0.79986 ±0.003\\
        sberbank-ai/ruBert-large                           & 0.80154 ±0.002\\
        IlyaGusev/mbart\_ru\_sum\_gazeta                   & 0.80566 ±0.001\\
        MoritzLaurer/mDeBERTa-v3-base-mnli-xnli            & 0.80710 ±0.001\\
        \textbf{DeepPavlov/xlm-roberta-large-en-ru-mnli}   & \bf{0.81708 ± 0.002} \\ \hline
        \textbf{Ensemble}                                  & \textbf{0.82995}\footnote{\href{https://www.kaggle.com/competitions/ruatd-2022-bi/leaderboard}{Kaggle private leaderboard link}}\\ \hline

    \end{tabular}
  \end{center}
  \caption{Results for binary classification for different models}
  \label{binaryTable}
\end{table}

\begin{table}[h!]
  \begin{center}
    \begin{tabular}{|l|c|c|}
      \hline \bf Model name & \bf Accuracy(validation) & \bf Accuracy(kaggle public/private) \\ \hline
        IlyaGusev/mbart\_ru\_sum\_gazeta                   & 0.6142 & 0.61459/0.61092 \\
        DeepPavlov/rubert-base-cased.                      & 0.6045 & 0.60433/0.60472 \\
        \textbf{DeepPavlov/xlm-roberta-large-en-ru-mnli}.  & \textbf{0.6242} & \textbf{0.62856/0.62644} \\ \hline

    \end{tabular}
  \end{center}
  \caption{Results for multiclass classification for different models}
  \label{multiclass_classification}
\end{table}

\section{Conclusion}
In this paper, we described our pipeline for the DIALOG-22 RuATD challenge. Our solution achieved 1st place in binary classification using ensembling techniques and 4th place for the multiclass classification task using only a single model. However, the proposed solution requires a lot of computational power, so it cannot be used in real-time systems to detect generated texts. But it gives us an understanding that we still need to upgrade methods to distinguish generated texts from human-written ones.

\bibliography{dialogue.bib}
\bibliographystyle{dialogue}

\end{document}